\documentclass[sigconf]{acmart}
\usepackage{fancyhdr}
\copyrightyear{2026}
\acmYear{2026}
\setcopyright{cc}
\setcctype{by-nc-nd}
\acmConference[WWW Companion '26]{Companion Proceedings of the ACM Web Conference 2026}{April 13--17, 2026}{Dubai, United Arab Emirates}
\acmBooktitle{Companion Proceedings of the ACM Web Conference 2026 (WWW Companion '26), April 13--17, 2026, Dubai, United Arab Emirates}
\acmPrice{}
\acmDOI{10.1145/3774905.3794688}
\acmISBN{979-8-4007-2308-7/2026/04}

\usepackage{booktabs}
\usepackage[table]{xcolor}
\usepackage{tabularray}
\UseTblrLibrary{booktabs}

\usepackage{algorithm}
\usepackage{algpseudocode}
\algnewcommand{\Parameters}[1]{\State \textbf{Parameters:} #1}

\AtBeginDocument{%
 }

\begin{document}

\title{Feedback Control for Small Budget Campaigns in Advertising}

\author{Sreeja Apparaju}
\orcid{0000-0001-6903-8937}
\affiliation{%
  \institution{Snap Inc.}
  \city{New York City}
  \country{United States}}
\email{sapparaju@snapchat.com}

\author{Yichuan Niu}
\orcid{0009-0000-5874-5783}
\affiliation{%
  \institution{Snap Inc.}
  \city{Palo Alto}
  \country{United States}}
\email{yniu@snapchat.com}

\author{Xixi Qi}
\orcid{0009-0005-6519-8710}
\affiliation{%
  \institution{Snap Inc.}
  \city{Palo Alto}
  \country{United States}}
\email{xqi@snapchat.com}

\begin{abstract}
Budget pacing is critical in online advertising to align spend with campaign goals under dynamic auctions. Existing pacing methods often rely on ad-hoc parameter tuning, which can be unstable and inefficient. We propose a principled controller that combines bucketized hysteresis with proportional feedback to provide stable and adaptive spend control. Our method provides a framework and analysis for parameter selection that enables accurate tracking of desired spend rates across campaigns. Experiments in real-world auctions demonstrate significant improvements in pacing accuracy and delivery consistency, reducing pacing error by 13\% and $\lambda$-volatility by 54\% compared to baseline method. By bridging control theory with advertising systems, our approach offers a scalable and reliable solution for budget pacing, with particular benefits for small-budget campaigns.
\end{abstract}



\maketitle

\section{Introduction}

In online digital advertising with real-time auctions, budget pacing is essential for ensuring that spending aligns with campaign goals. Advertisers may wish to distribute  budgets evenly over time or follow traffic patterns, requiring pacing systems that dynamically adjust spending velocity. A pacing system sets a desired spending plan, monitors actual spend, and continuously corrects deviations in near real time. By minimizing the gap between the desired and actual spending velocity, advertisers achieve predictable delivery and efficient budget use.

Designing such systems is challenging. Auction dynamics, stochastic traffic, noisy signals, and changing market conditions can all destabilize pacing. In practice, pacing control is often based on empirical tuning, which provides neither stability nor efficiency guarantees.

In this work, we design a controller that combines hysteresis control with proportional feedback, offering a principled alternative to trial-and-error methods. Our contributions are threefold:
\begin{enumerate}
    \item We implement a bucketized hysteresis controller to improve stability and dynamics, and compare its performance with a legacy variable-step pacing system.
    \item We provide an analytical framework to guide parameter selection, enabling personalized tracking of spend rates.
    \item We validate our approach through real-world auction experiments to demonstrate significant improvements over legacy feedback-based pacing.
\end{enumerate}

By bridging control theory with modern advertising systems, we provide a scalable and systematic approach to budget pacing, achieving higher accuracy for small-budget campaigns while remaining broadly applicable across use cases.

\section{Related Work}

Budget pacing in online advertising has been studied extensively in both the machine learning and control theory communities. Early work focused on optimization-based formulations of budget allocation and pacing. Feldman et al.~\cite{feldman2006budget} addressed budget optimization in search advertising auctions, while Agarwal et al.~\cite{agarwal2014budget} and Xu et al.~\cite{xu2015smart} developed practical pacing algorithms demonstrating the importance of balancing spend delivery with campaign effectiveness. More recently, Nguyen et al.~\cite{nguyen2023practical} presented simulation-based evaluation for sponsored search pacing at eBay.

A complementary line of research applies feedback control theory in online advertising. Karlsson and collaborators~\cite{karlsson2013applications, karlsson2021scalable, karlsson2022feedback, karlsson2022multiobjective, karlsson2024optimization} explored multi-objective optimization, adaptive rate and price control, and revenue-profit trade-offs, highlighting the flexibility of control-based approaches. Zhang et al.~\cite{zhang2016feedback} applied classical feedback control to real-time display advertising, while Aghdaii~\cite{aghdaii2020adsopt, aghdaii2020auction} provided accessible overviews of pacing and auction optimization. Together, these works demonstrate the potential of control-theoretic methods to stabilize delivery in dynamic auction environments.

Our work is most closely related to this feedback control stream. However, prior approaches typically rely on proportional or integral adjustments, which can suffer from oscillations or require delicate parameter tuning. In contrast, we propose a \emph{Bucketized Hysteresis Controller} that discretizes the error space into multiple bands and applies multiplicative updates in variable steps. This strategy is inspired by classical bang-bang control but extended with quantized step sizes, analogous to gain scheduling, to ensure both rapid convergence and stability. To our knowledge, this is the first application of multi-step hysteresis control for budget pacing in real-time auctions, bridging classical control techniques with the demands of modern advertising platforms.

\section{System Setup}
\subsection{Bidding and Auction}
In an advertising system, the probability of conversion is often used to represent the likelihood of a desired event. In this work, we use the probability of an ad impression being seen by users as the conversion event. Assuming $p(event)$ is perfectly calibrated, an ad will eventually receive a click when the cumulative sum of $p(event)$ across multiple impressions reaches 1.0. To participate in an auction, advertisers must place a bid, \emph{max bid}, representing the amount they are willing to pay for an event. 

For a single impression opportunity, each ad submits its final bid as follows ~\cite{aghdaii2020auction}: 
\begin{equation}
\text{final\_bid} = \text{max\_bid} \times p(\text{event})
\end{equation}
The ad with the highest bid is shown to the user, and the advertiser is charged either the final bid in a first-price auction or the runner-up’s bid in a second-price auction. At this stage, the revenue system is \emph{not paced} because total ad spend is entirely dependent on auction traffic. In practice, however, advertisers operate within daily or lifetime budget constraints, ensuring that cumulative spending on all winning bids does not exceed these limits.

\subsection{Pacing}
To enforce such constraints, a pacing control variable is introduced into the final bid calculation ~\cite{aghdaii2020adsopt}: 
\begin{equation}
\text{paced\_final\_bid} = \lambda \times \text{final\_bid}
\end{equation}

where $\lambda \in (0,1]$ acts as a bid modifier. This type of pacing is also known as discount pacing. The final auction and pricing process are then governed by the $\lambda$ stochastically. From bidding and auction to actual budget spending, the pacing control variable $\lambda$ influences the budget spent by rescaling bids, thereby increasing or decreasing the win rate in auctions. 

In most cases, $\lambda$ updates are scheduled at fixed intervals using a cron job. Under ideal conditions $\lambda$ remains constant as long as auction intensity and the total number of impression opportunities remain unchanged. However, in practice, $\lambda$ must adapt to environmental changes to ensure that the remaining budget is spent appropriately throughout the day or campaign lifetime, following the prescribed budget spending plan. The ultimate objective of this work is to design $\lambda$ controllers to achieve precise and robust pacing.

\paragraph{Pacing for Small-Budget Campaigns}
Small-budget campaigns exhibit fundamentally different pacing dynamics compared to large-budget campaigns. They have limited spending capacity spread across many auction opportunities, hence even minor adjustments to $\lambda$ produce disproportionately large changes in delivery rate. A marginal increase in $\lambda$ can lead to an immediate and exhaustive surge in spend, while a marginal decrease can halt delivery entirely. In this high-gain regime, these instability mechanisms are particularly pronounced for small-budget ad lines with broad targeting, where auction frequency amplifies the sensitivity of spend to bid scaling. This motivates the use of nonlinear control strategies that explicitly limit effective loop gain near the target operating point, as pursued in this work.

\section{Pacing Controller}

The production pacing controller currently deployed in practice serves as the baseline for our proposed method.  

\paragraph{Lambda Update}
The production system employs a variable step size controller with multiplicative updates to the pacing parameter $\lambda$, which scales to align observed delivery with a predefined target rate. The core control law is a multiplicative update rule (Algorithm~\ref{alg:prod_controller}), and the magnitude of the adjustment is governed by a scale $\alpha$ which acts as the effective learning rate of the controller. 

\paragraph{Scale Selection}

The scale $\alpha_t$ is adapted based on recent pacing history, we evaluate it over a short lookback window of $\lambda \times$ bid values. From this trajectory, a \emph{fluctuation factor} $F$ is computed, which measures the degree of oscillation in the system. The fluctuation factor quantifies the ratio of cumulative variation to net displacement in recent updates. This distinguishes smooth, consistent trends ($F \approx 1$) from oscillatory or unstable dynamics ($F \gg 1$).

\begin{equation}
F = \frac{\sum_{i=1}^{n-1} \left| x_{i+1} - x_i \right|}{\left| x_n - x_1 \right|},
\end{equation}

The controller applies scale updates as discrete and fixed increments or decrements according to predefined system parameters, ensuring stable response. The step size is not estimated continuously; instead it is \textbf{quantized into upward and downward adjustments of fixed magnitude}, applied conditionally on recent stability.
\vspace{3mm}
\begin{algorithm}[t]
\caption{Variable-Step Multiplicative Pacing Controller (Baseline)}
\label{alg:prod_controller}
\begin{algorithmic}[1]
\Require $\lambda_t$, step size $\alpha_t$, observed rate $\hat{r}_t$, desired rate $r_t$, recent trajectory $\{x_1,\dots,x_n\}$ 
\Parameters Upward rate $\eta_{+}$, downward rate $\eta_{-}$, threshold $\tau$, tolerance $\varepsilon>0$; optional bounds $\alpha_{\min},\alpha_{\max}$
\Ensure Updated $(\lambda_{t+1}, \alpha_{t+1})$
\Statex \textbf{// Scale (step size) adaptation}
\If{$n < 2$}
    \State $\alpha_{t+1} \gets \alpha_t$ \Comment{insufficient history}
\Else
    \State $\text{displacement} \gets |x_n - x_1|$
    \State $\text{distance} \gets \sum_{i=1}^{n-1} |x_{i+1} - x_i|$
    \If{$\text{displacement} = 0$}
        \State $F \gets \infty$
    \Else
        \State $F \gets \text{distance} / \text{displacement}$
    \EndIf
    \If{$F \le 1$}
        \State $\alpha_{t+1} \gets \alpha_t (1 + \eta_{+})$ \Comment{stable trend \(\Rightarrow\) speed up}
    \ElsIf{$F > \tau$}
        \State $\alpha_{t+1} \gets \alpha_t (1 - \eta_{-})$ \Comment{oscillatory \(\Rightarrow\) slow down}
    \Else
        \State $\alpha_{t+1} \gets \alpha_t$ \Comment{no change}
    \EndIf
\EndIf
\If{bounds defined}
    \State $\alpha_{t+1} \gets \min\{\alpha_{\max},\, \max\{\alpha_{\min},\, \alpha_{t+1}\}\}$ \Comment{clamp (optional)}
\EndIf
\Statex \textbf{// Multiplicative \(\lambda\) update}
\If{$|\hat{r}_t - r_t| \le \varepsilon$}
    \State $\lambda_{t+1} \gets \lambda_t$ \Comment{no update}
\ElsIf{$\hat{r}_t < r_t$}
    \State $\lambda_{t+1} \gets \lambda_t \,(1 + \alpha_{t+1})$ \Comment{under-delivery}
\Else
    \State $\lambda_{t+1} \gets \lambda_t \,(1 - \alpha_{t+1})$ \Comment{over-delivery}
\EndIf
\State \Return $(\lambda_{t+1}, \alpha_{t+1})$
\end{algorithmic}
\end{algorithm}
\vspace{-4mm}

\section{Bucketized Hysteresis Controller}
We propose a Bucketized Hysteresis Controller (BHC), a novel nonlinear control strategy that addresses the limitations of reactive pacing methods based on incremental updates. Inspired by classical bang-bang control, we adapt this paradigm to pacing systems, where oscillations arise from overspending and efficiency fluctuations rather than actuator constraints.

\subsection{Controller Design}
The BHC introduces variable adjustment steps, creating a multi-step, quantized controller. The step sizes are determined by the system error, following a principle analogous to gain scheduling, in which controller parameters adapt to the operating point. Large predefined steps correct large errors to accelerate convergence, while smaller steps correct minor errors to preserve stability.

\subsection{Controller Methodology}

\noindent At each time step $t$, the controller proceeds in three stages:   \paragraph{Error Calculation:} The normalized error between the desired rate $\theta_d$ and the observed rate $\theta_o$ is computed as,
\begin{equation}
E_t = \frac{\theta_d - \theta_o}{\theta_d}, \qquad u_t = \operatorname{sgn}(E_t),
\end{equation}
where $\theta_d$ is the desired rate and $\theta_o$ is the observed rate. The sign $u_t$ sets the adjustment direction.  \paragraph{Error Discretization} The error magnitude is mapped into one of $K$ non-overlapping bands with thresholds $\{\tau_i\}_{i=1}^K$ and associated gains $\{s_i\}_{i=1}^K$. The active band is selected as
\begin{equation}
k = \max\{i \mid \tau_i \le |E_t|\}
\end{equation}
\paragraph{Lambda Update}: The pacing multiplier is then updated using the selected gain and direction,
\begin{equation}
\lambda_t = \lambda_{t-1}\bigl(1 + s_k\,u_t\bigr).
\end{equation}
This procedure produces multiplicative updates whose magnitude depends on the error band, balancing rapid convergence with long-term stability.
\begin{algorithm}[h]
\caption{Bucketized Hysteresis Update}
\label{alg:bucket_hysteresis}
\begin{algorithmic}[1]
\Require Current pacing multiplier $\lambda_t$, observed rate $o_t$, desired rate $d_t$, thresholds $\{T_1,\ldots,T_K\}$, scales $\{s_1,\ldots,s_K\}$, tolerance $\varepsilon>0$
\Ensure Updated $\lambda_{t+1}$
\State $E_t \gets \dfrac{d_t - o_t}{d_t}$ \Comment{relative error}
\If{$\lvert o_t - d_t \rvert < \varepsilon$}
    \State $\lambda_{t+1} \gets \lambda_t$ \Comment{within tolerance: no update}
\Else
    \If{$o_t < d_t$}
        \State $dir \gets +1$ \Comment{under-delivery: upward adjustment}
    \Else
        \State $dir \gets -1$ \Comment{over-delivery: downward adjustment}
    \EndIf
    \State $j \leftarrow \max\{i | T_i \le |E_t|\}$ \Comment{select band index}
    \State $\lambda_{t+1} \gets \lambda_t \cdot \bigl(1 + s_j \cdot dir\bigr)$ \Comment{multiplicative update}
\EndIf
\State \Return $\lambda_{t+1}$
\end{algorithmic}
\end{algorithm}

\subsection{Band Threshold and Gain Selection}
BHC parameters are derived systematically through empirical system identification. We analyze historical pacing trajectories from small-budget campaigns optimizing
for the same delivery objectives to characterize the relationship between the
normalized pacing error $E_t$ and subsequent changes in the pacing multiplier
$\lambda$ over short forward-looking windows.

Band thresholds $\{\tau_i\}$ are selected to partition the normalized error space into coarse control regimes (e.g., small, moderate, and large deviations), while the gains $\{s_i\}$ for each band are set proportional to the mean predicted $\lambda$ adjustment associated with that error range. Importantly, the resulting thresholds and gains are global controller parameters, reused across campaigns and fixed during online deployment.

\section{Online Experiments}

\paragraph{Setup.}
We evaluate the proposed controller using live experiments within a designated Budget Group Identifier (BGID), an internal mechanism that partitions traffic to enable controlled experimentation and fine-grained measurement. Our evaluation focuses on ad lines with small budgets, where delivery is highly sensitive to fluctuations. Small budgets with large targeting groups increases auction frequency and amplify the sensitivity of the relationship between $\lambda$ and the spend rate. 

\paragraph{Metrics.}

We evaluate our approach using the following three metrics: 
\begin{enumerate}
    \item \emph {System-Level Pacing Error} (PE) measuring day-long smoothness, $\mathrm{PE}=\tfrac{1}{N}\sum_{j=1}^{N}\tfrac{\lvert p^S_j-p^T_j\rvert}{p^T_j}$~\cite{nguyen2023practical}; where $p^S_j$ and $p^T_j$ are actual and target spends
    \item  \emph{$\lambda$-volatility} measuring the coefficient of variation, $\mathrm{CV}_\lambda=\operatorname{std}(\lambda_{1:T})/\operatorname{mean}(\lambda_{1:T})$; 
    \item \emph{CPM} measuring cost for the advertiser to show their ad to users. 
\end{enumerate}

\begin{figure*}[ht!]
\centering
\includegraphics[alt={A 5x2 grid of plots. The left column shows spend accumulation and the right shows lambda trends for five experiments: BHC, AOF, ALU, RUDM, and SSDM. In each case, the test group is compared against a control group and an ideal spend line.},width=10cm, 
  height=12cm ]{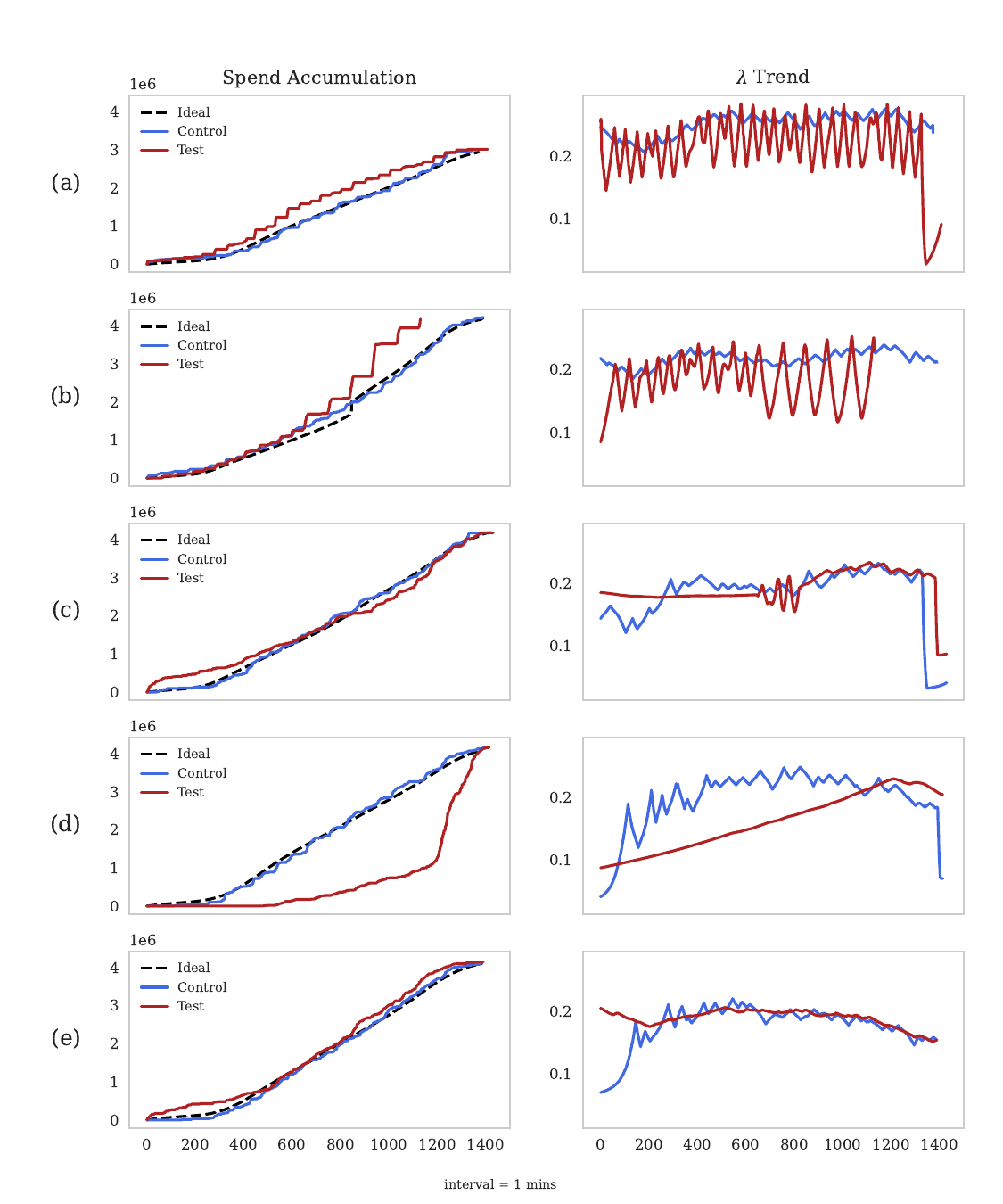}

\caption{For each strategy (a-e), the left panel shows the daily cumulative spend of our Test controller (red) against the Production controller (blue) and an Ideal target (black, dashed). The right panel shows the corresponding controller output, \(\lambda\).}

\label{fig:spend_lambda_combined}
\end{figure*}

\subsection{Online Testing}
Our evaluation compares the proposed BHC variants against the production pacing controller currently deployed at scale. The production controller incorporates adaptive feedback mechanisms and represents a mature, carefully engineered baseline. We discuss the resulting constraints on baseline selection and experimental scope in Section~\ref{sec:limitations}.

\paragraph{Prod vs. Standard Bucketized Hysteresis Configuration (BHC)}
We first replaced the production controller’s fixed increment update with a BHC formulation using a standard band configuration, i.e., default thresholds $\{T_i\}_{i=1}^{K}$ and scales $\{s_i\}_{i=1}^{K}$ originally tuned for small-budget traffic. While this change increased sensitivity to pacing error, online tests revealed pronounced oscillations in controller output and unstable spend trajectories, rendering this configuration unsuitable at scale.

\paragraph{Prod vs. Averaging Observed Feedback (AOF)}
To reduce the observed oscillations, we introduced temporal averaging of the observed spend signal over $M{=}20$ control cycles,
\[\bar r_t = \tfrac{1}{M} \sum_{i=0}^{M-1} \hat r_{t-i}.\]
This modification substantially smoothed the input signal but resulted in delayed corrections and coarse adjustment behavior, degrading overall pacing performance.

\paragraph{Prod vs. Averaging \texorpdfstring{$\lambda$}{lambda} Updates (ALU)}
Instead of filtering the observed feedback, we next applied averaging directly to the controller output. In this variant, $\lambda$ was updated each cycle but averaged over $L{=}10$ iterations before application:
\[\lambda_t = \tfrac{1}{L} \sum_{i=0}^{L-1} \tilde\lambda_{t-i}.\]
This approach produced stable spend trajectories and significantly reduced variability in $\lambda$, while preserving responsiveness to pacing error.

\paragraph{Prod vs. Slowed Bands}
Finally, we reduced the gain of the initial BHC bands to further stabilize the controller during ramp-up. This conservative tuning produced two qualitatively distinct operating regimes. During the initial transient, \textit{Ramping State (RSDM)} the reduced loop gain yields an overdamped response with slow convergence, requiring nearly half a day to converge. Once the tracking error falls within the deadband and the system settles, it enters the \textit{Steady State (SSDM)}

\begin{table}[ht!]
    
    \caption{Relative performance of BHC variants compared to the production baseline. Values show percentage change: negative values indicate improvement (lower error and volatility).}
    
    \label{tab:results}
    \begin{tblr}{
      colspec = {@{} l | c c c c @{}},
      row{odd[3-7]} = {bg=gray!15}, 
      row{1} = {font=\bfseries}, 
    } 
    \toprule
    & BHC & AOF & ALU & SSDM \\
    \midrule 
    PE & +88.23\% & +173.12\% & +69.94\% & \textbf{-13.06\% }\\
    $\lambda$-volatility & +132.93\% & +210.22\% & \underline{-2.63\%} & \textbf{-53.78\%} \\
    \midrule 
    CPM & +1.4\% & +2.3\% & \textbf{-1.64\%} & \underline{-1.07\%} \\
    \bottomrule
    \end{tblr}
\end{table}

\subsection{Analysis}

\paragraph{Prod vs BHC} 
In Figure~\ref{fig:spend_lambda_combined}(a), the Test controller shows persistent oscillations in $\lambda$ (right panel) due to increased sensitivity to error. These oscillations drive unstable spending in the cumulative spend curve (left panel), where delivery surges above the Ideal (black dashed line) before collapsing and resettling. This pattern reflects excessive loop gain, producing a limit cycle instead of convergence. The high PE and $\lambda$-volatility in Table~\ref{tab:results} confirms the instability and highlights the large fluctuations.

\paragraph{Prod vs AOF} 
Figure~\ref{fig:spend_lambda_combined}(b) shows a staircase-like spending cycle (left panel), caused by delayed error corrections introduced by averaging observed spend. The $\lambda$-trend (right panel) makes coarse, stepwise adjustments in Test controller (red) instead of tracking smoothly, which indicates sluggish adaptation. Although this filter reduces short-term noise, it introduces excessive lag, making the system unable to respond promptly to deviations. This is reflected in both highest PE (+173.12\%) and $\lambda$-volatility (+210.22\%) from Table~\ref{tab:results}, identifying AOF as the \emph{worst performing controller variant}.

\paragraph{Prod vs ALU} 
In Figure~\ref{fig:spend_lambda_combined}(c), the ALU controller achieves smooth, continuous spending that closely follows the ideal trajectory (black dashed line). The corresponding $\lambda$-trend (right panel) shows moderate adjustments, where averaging dampens high-frequency noise without sacrificing responsiveness. Importantly, ALU delivers the lowest CPM (-1.64\%) among all controllers, demonstrating efficiency gains from stabilized pacing. 

\paragraph{Prod vs Slowed Bands}
Figures~\ref{fig:spend_lambda_combined}(d) and~\ref{fig:spend_lambda_combined}(e) illustrate the two damping regimes. Figure~\ref{fig:spend_lambda_combined}(d) captures behavior during the first 12 hours (RSDM), where the conservative gain scales result in slow, cautious adjustments to $\lambda$ (right panel). This produces visible underspending in Test (red) relative to both Production (blue) and the Ideal trajectory 
(black dashed) as the controller gradually ramps up. The small $\lambda$ 
increments prevent the overshoot observed in standard BHC but delay convergence.

Figure~\ref{fig:spend_lambda_combined}(e) shows the same controller after convergence (SSDM). Here, the identical conservative tuning that caused slow ramp-up now provides excellent stability: $\lambda$ exhibits minimal fluctuations (right panel) and spending tracks the ideal trajectory smoothly (left panel). Table~\ref{tab:results} reports aggregate metrics across the full 24-hour period: PE improves by 13.06\%, $\lambda$-volatility drops by 53.78\%, and CPM decreases modestly by 1.07\%.

Thus, RSDM represents a necessary transient for safe convergence, while SSDM
represents the intended long-term operating regime.

\section{Conclusion}

Our findings are consistent with foundational control theory while addressing 
the practical challenges of online advertising. The experiments highlight a 
fundamental trade-off between fast transient response and steady-state stability. 
A high-gain controller enables rapid tracking of the setpoint but risks instability 
and oscillations, as demonstrated by the standard BHC variant. Conversely, a 
low-gain controller ensures stability and minimizes steady-state error but suffers 
from sluggish transient response and slow disturbance rejection.

Among the variants tested, the slowed-bands (SB) controller achieved the best 
stability metrics, reducing pacing error by 13\% and $\lambda$-volatility by 54\% 
compared to production. This conservative tuning eliminates oscillations and 
provides smooth, predictable spending once the system converges to steady state 
(SSDM). However, the same low-gain design that ensures stability also causes 
slow initial convergence (RSDM), requiring approximately 12 hours to reach target 
spend rates. For small-budget campaigns—where high sensitivity amplifies the 
impact of control actions—this trade-off proves worthwhile, as the extended 
convergence period is tolerable relative to the campaign lifetime, while the 
steady-state stability is critical for consistent delivery.

By bridging control theory with the demands of real-world advertising systems, 
our bucketized hysteresis framework with adaptive gain tuning provides a scalable 
and principled approach to budget pacing. The slowed-bands variant demonstrates 
that systematic parameter selection based on control-theoretic principles can 
outperform ad-hoc tuning methods, particularly for the challenging regime of 
small-budget, high-sensitivity campaigns.

\section{Future Work}

These results motivate the development of an adaptive pacing strategy. A promising direction is a multi-modal controller bands with predefined operating modes (e.g., “Turbo,” “Regular,” “Fine”), selected dynamically according to the magnitude of the tracking error. Large, persistent errors would trigger a high-gain Turbo mode for fast convergence, while small errors would engage a low-gain Fine mode to ensure stable tracking. Future work will focus on designing this adaptive mechanism and ensuring robustness to external disturbances, such as changes in bid strategy and updates to core ranking models.

\section{Limitations}
\label{sec:limitations}

Evaluating pacing algorithms in live production environments is subject to significant 
practical constraints that are standard across industrial advertising research 
\cite{nguyen2023practical, karlsson2021scalable}. Specifically: (1) implementing 
alternative baseline methods requires substantial re-engineering of core auction 
infrastructure and bidding pipelines; (2) deploying unvalidated pacing strategies 
poses direct revenue risk to active advertiser campaigns and must pass rigorous 
internal review; and (3) proprietary platform-specific features in many published 
methods \cite{nguyen2023practical, xu2015smart} cannot be directly replicated in 
our environment. These constraints necessitate that online experiments be limited 
to methods with established internal validation paths.
\bibliographystyle{ACM-Reference-Format}
\bibliography{base}


\end{document}